\documentclass{article}

\usepackage{scml2026}

\usepackage{multicol}
\usepackage[utf8]{inputenc} 
\usepackage[T1]{fontenc}    
\usepackage{hyperref}       
\usepackage{url}            
\usepackage{booktabs}       
\usepackage{amsfonts}       
\usepackage{amsmath}
\usepackage{amssymb}
\usepackage{nicefrac}       
\usepackage{microtype}      
\usepackage{cleveref}       
\usepackage{graphicx}
\usepackage{natbib}
\usepackage{doi}
\usepackage{wrapfig}
\usepackage{float}

\title{Cluster-Weighted EDMD}
\renewcommand{\shorttitle}{CW-EDMD}

\date{}

\author{%
	Lorenzo Tomaz\thanks{Corresponding author.}, Judd Rosenblatt, Flavio Kicis, Thomas B. Jones, Diogo Schwerz de Lucena \\
	AE Studio \\
	\texttt{\{lorenzo, judd, flavio.kicis, thomas, diogo\}@ae.studio}
}

\hypersetup{
pdftitle={Cluster-Weighted EDMD},
pdfsubject={SCML2026 Abstract},
pdfauthor={Lorenzo Tomaz, Judd Rosenblatt, Flavio Kicis, Thomas B. Jones, Diogo Schwerz de Lucena},
pdfkeywords={Koopman operator, EDMD, cluster-weighted models, EM, dynamical systems},
}

\begin{document}
\maketitle


\begin{multicols}{2}

Extended Dynamic Mode Decomposition (EDMD) is the canonical data-driven approximation of the Koopman operator~\citep{williams2015edmd,mezic2005spectral,brunton2022modern,korda2018mpc,mauroy2020koopman}. Prior partitioning approaches predefine the partition by basin label~\citep{williams2015edmd}, phase-space stitching~\citep{sinha2020phase}, or operating-regime indicator~\citep{peitz2019switched}. A complementary line of work enriches the global observable basis through learned dictionaries~\citep{li2017dictionary}, deep autoencoders~\citep{lusch2018deep}, or kernel methods~\citep{williams2015kernel}, orthogonal to the partitioning direction pursued here. We introduce \emph{Cluster-Weighted EDMD} (CW-EDMD), which learns the partition jointly with per-cluster operators via Expectation-Maximization (EM) on a cluster-weighted-model joint density~\citep{gershenfeld1999nature,ingrassia2014linear,punzo2014polynomial}, with responsibilities proportional to the product of geometric proximity and per-cluster prediction accuracy. Across three classical systems (36 configurations, 10 seeds), CW-EDMD improves over EDMD at the matched polynomial lift degree, including where EDMD itself saturates.

\noindent\textbf{Method.} CW-EDMD fits a separate Koopman operator per cluster via EM (full derivation in Appendix~A), with responsibilities combining geometric proximity and per-cluster prediction residual. Each cluster has a center, a covariance, and a Koopman matrix fit on a recentered polynomial lift of degree $q$. The key departure from a standard Gaussian mixture is residual-awareness: a cluster earns responsibility for a training transition in proportion to both how close the current state is to its center \emph{and} how accurately it predicts the next state, so the partition tracks where each operator predicts well rather than where data is dense. Given responsibilities, each Koopman matrix is updated in closed form by responsibility-weighted least squares, generalizing the standard EDMD solution to the per-cluster regime.

\textbf{Experimental setup.} We evaluate on three classical systems: the Lorenz attractor, a damped pendulum (non-polynomial $\sin\theta$ RHS), and a double-well Duffing oscillator. Each system is swept across 12 configurations varying sampling distribution, data size, domain, integrator step, and fit budget, with 10 fixed seeds per configuration (full details in Appendix~B). Matched-degree: CW-EDMD-$(q,G)$ vs.\ EDMD-$q$ on 10 paired seeds; per-seed metric is mean $\ell_2$ test error (one-step or 5\,s rollout, separate cells). A cell is a \emph{win} (W) if paired Wilcoxon gives $p<0.05$ with lower CW-EDMD across-seed mean, a \emph{loss} (L) if higher, a \emph{tie} (T) otherwise.

\begin{table}[H]
\caption{Matched-degree CW-EDMD vs.\ EDMD, paired-Wilcoxon W/L/T and median error ratio (EDMD / CW-EDMD; ratio $>1$ favors CW-EDMD), aggregated over all CW-EDMD $(q, G)$ variants and all 12 configurations per system, at one-step and 5\,s rollout.}
\label{tab:headline}
\centering\footnotesize
\setlength{\tabcolsep}{3pt}
\begin{tabular}{lcccc}
\toprule
& \multicolumn{2}{c}{W / L / T} & \multicolumn{2}{c}{ratio} \\
\cmidrule(lr){2-3}\cmidrule(lr){4-5}
System & 1-step & 5\,s & 1-step & 5\,s \\
\midrule
Pendulum & 33/0/3 & 32/0/4 & 57$\times$ & 55$\times$ \\
Duffing & 35/0/1 & 35/0/1 & 2.7$\times$ & 2.5$\times$ \\
Lorenz & 65/4/3 & 58/0/14 & 12$\times$ & 8.3$\times$ \\
\bottomrule
\end{tabular}

\end{table}

\noindent\textbf{Results.} CW-EDMD outperforms EDMD at the matched polynomial lift on all three systems (Table~\ref{tab:headline}; accuracy-parameter tradeoffs in Figures~\ref{fig:pareto-duffing}--\ref{fig:pareto-lorenz}). Across the $288$ paired tests in Table~\ref{tab:headline}, CW-EDMD records $258$ wins, $4$ losses, and $26$ ties; all $4$ losses are the smallest-$N$ Lorenz configuration (Appendix~C). Disabling the residual factor in the E-step (Appendix~D) splits the gain: residual-awareness carries it on the pendulum and at low $q$ on Lorenz/Duffing; at the $q$ where EDMD saturates, geometry-only partitioning suffices.

\section*{Acknowledgments}
This work was funded by AE Studio and the AI Alignment Foundation. The authors thank colleagues for discussions and feedback. Code, configs, and per-seed results: \url{https://github.com/agencyenterprise/cluster_weighted_edmd}.

\bibliographystyle{plainnat}
\bibliography{references}

\end{multicols}

\raggedbottom

\section*{Appendix A: Full method derivation}

\paragraph{Related work and positioning.} The idea of partitioning phase space and fitting a separate Koopman operator per region is not new. The original EDMD paper~\citep{williams2015edmd} demonstrates a partitioned EDMD on the Duffing oscillator by identifying the basins of attraction from a leading Koopman eigenfunction and fitting separate operators to each basin, an early observation that per-region operators on dynamically meaningful partitions can outperform a single Koopman operator on the full phase space. Nandanoori, Sinha, and Yeung~\citep{sinha2020phase} formalize phase-space stitching of region-specific Koopman operators for large-scale dynamical systems. Peitz and Klus~\citep{peitz2019switched} develop a switched/multi-model Koopman formulation in which different operating regimes are assigned distinct linear Koopman models, with regime indicators driving the switch. Earlier mixture-of-experts work in the statistics literature~\citep{jordan1994hme,gershenfeld1999nature,ingrassia2014linear,punzo2014polynomial} provides the joint-density EM machinery that CW-EDMD reuses. The contribution of CW-EDMD relative to these works is the combination of three elements: (i) the partition is \emph{learned jointly} with the per-cluster operators by EM rather than predefined by basin label, geometry, or operating-regime indicator; (ii) the responsibilities used in the expectation step combine geometric proximity with per-cluster prediction residual, so that the partition tracks where each operator predicts well rather than where data is geometrically dense; and (iii) the per-cluster predictor is the discrete EDMD operator on a recentered polynomial lift, giving a clean apples-to-apples baseline of CW-EDMD against EDMD at matched lift degree. The geometry-only ablation in Appendix~D shows, on Duffing, that element (ii) is what drives the matched-degree advantage.

\paragraph{Two orthogonal axes of Koopman approximation.} Beyond the partitioning literature surveyed above, a parallel line of work improves Koopman approximation by enriching the global observable basis: dictionary learning~\citep{li2017dictionary}, deep Koopman autoencoders~\citep{lusch2018deep,takeishi2017koopman,yeung2019deep}, and kernel EDMD~\citep{williams2015kernel}. These methods retain a single global operator on a richer learned or kernel basis. CW-EDMD makes the orthogonal bet: a partitioned state space in which a simple local basis (recentered monomials) suffices per region. The two directions address different inductive failure modes --- basis insufficiency (e.g., $\sin\theta$ has no exact polynomial expansion at any finite degree) versus operator insufficiency (e.g., multi-attractor dynamics that no single linear map can represent) --- and compose naturally: a learned per-cluster basis is the obvious next step. For the matched-degree comparisons reported here, our baseline is therefore vanilla EDMD at the same monomial lift, which isolates the partitioning contribution; head-to-head comparison against learned-basis methods is left for future work, where per-cluster learned bases are the natural target.

\paragraph{CWM joint density.} We now define the model formally. Given $N$ paired training transitions $\{(x_t^{(i)}, x_{t+1}^{(i)})\}_{i=1}^{N}$ in $\mathbb{R}^d$, we model the joint density over consecutive states as a mixture of $G$ components,
\[
p(x_t, x_{t+1}) = \sum_{g=1}^{G} \pi_g\, p_X(x_t \mid g)\, p_{Y\mid X}(x_{t+1} \mid x_t, g),
\]
where $\pi_g \geq 0$ are mixture weights summing to one. The two factors per component have distinct roles. The first factor $p_X(x_t \mid g) = \mathcal{N}(x_t;\, c_g, \Sigma_g)$ is a Gaussian centered at $c_g$ with covariance $\Sigma_g$; it measures geometric proximity of the current state to cluster $g$. The second factor $p_{Y\mid X}(x_{t+1} \mid x_t, g) = \mathcal{N}(\Delta x_g;\, 0, \sigma_g^2 I)$ is a Gaussian on the prediction residual $\Delta x_g = x_{t+1} - \hat{x}_{t+1\mid g}$, where $\hat{x}_{t+1\mid g}$ is cluster $g$'s prediction of the next state (defined in the following paragraph); it measures how accurately cluster $g$ predicts the observed transition. During training, $\hat{x}_{t+1\mid g}$ is computed from the current iterate of $K_g$ at each E-step; at inference, $x_{t+1}$ is unavailable and cluster selection reverts to geometric proximity alone. The train time residual factor is what distinguishes CW-EDMD from a standard Gaussian mixture model (GMM): during training, a GMM assigns cluster responsibilities based on geometric proximity alone, whereas CW-EDMD additionally requires a cluster to predict the observed transition well, so the learned partition reflects predictive accuracy rather than data density.

\paragraph{Per-cluster Koopman operator.} For each cluster $g$ with center $c_g$, we lift the recentered state $x_t - c_g$ into a higher-dimensional feature space via a monomial map $\Phi: \mathbb{R}^d \to \mathbb{R}^{M_q}$, where $\Phi$ collects all monomials up to total degree $q$ and $M_q = \binom{d+q}{q}$ is the number of such monomials. The first entry of $\Phi$ is the constant $1$, followed by the $d$ linear monomials, then quadratics, and so on. The per-cluster Koopman matrix $K_g \in \mathbb{R}^{M_q \times M_q}$ propagates the lifted state forward one step, and the predicted next state is recovered by projecting back to $\mathbb{R}^d$ via $P_d$, the matrix that selects the $d$ linear-monomial entries of the lift:
\[
\hat{x}_{t+1\mid g} = c_g + P_d\, K_g\, \Phi(x_t - c_g).
\]
We fit the full square $K_g$ rather than a rectangular matrix mapping directly to the state for the standard reason: the square form is the EDMD discretization of the Koopman operator on the full lifted space~\citep{williams2015edmd,brunton2022modern}, preserving the lifted-space spectrum for eigenfunction extraction and downstream control. The projection $P_d$ is applied only at prediction time, so the optimization targets the full $M_q$-dimensional lifted residual; the cost is $M_q^2 - d\,M_q$ unused parameters per cluster, reported in all parameter counts.

\paragraph{EM updates.}\label{par:em-updates} EM alternates between two steps.

The \textbf{E-step} computes, for every training transition, a soft responsibility score for each cluster. The responsibility of cluster $g$ for transition $i$ is proportional to the product of two terms: how likely the current state $x_t^{(i)}$ is under the cluster's geometric Gaussian, and how likely the observed next state is under the cluster's prediction residual Gaussian. Concretely,
\[
r_{ig} \propto \pi_g\, \mathcal{N}(x_t^{(i)}; c_g, \Sigma_g)\, \mathcal{N}\!\bigl(\Delta x_g^{(i)}; 0, \sigma_g^2 I\bigr).
\]

The \textbf{M-step} updates each cluster's parameters using responsibility-weighted averages: $c_g$, $\Sigma_g$, $K_g$, $\sigma_g^2$, and the mixture weight $\pi_g = (\sum_i r_{ig})/N$ --- the responsibility-weighted fraction of transitions assigned to cluster $g$. The cluster center is updated to the responsibility-weighted mean of the current states assigned to it, and the covariance to the corresponding responsibility-weighted scatter matrix:
\[
c_g = \frac{\sum_i r_{ig}\, x_t^{(i)}}{\sum_i r_{ig}}, \qquad
\Sigma_g = \frac{\sum_i r_{ig}\, (x_t^{(i)} - c_g)(x_t^{(i)} - c_g)^\top}{\sum_i r_{ig}}.
\]
The formulas above are the maximum-likelihood limit. In implementation we place weak conjugate priors on each cluster's parameters to stabilize the M-step on small clusters and enable empty-cluster pruning (next paragraph): a Gaussian prior $c_g \sim \mathcal{N}(\mu_0,\, \Lambda_0^{-1})$ on the center, an Inverse-Wishart prior $\Sigma_g \sim \mathcal{W}^{-1}(\Psi_0,\, \nu_0)$ on the covariance, and a symmetric Dirichlet prior $\boldsymbol{\pi} \sim \mathrm{Dir}(\alpha_0)$ on the mixture weights. The M-step is the corresponding MAP update, which interpolates each prior with the responsibility-weighted data statistics and recovers the formulas above as the cluster responsibility mass $R_g = \sum_i r_{ig}$ grows. We use $\mu_0 = \bar{x}$ (training mean), $\Lambda_0 = 10^{-2}\, I$, $\Psi_0 = \psi_0\, I$, $\nu_0 = d + 2$, and $\alpha_0 = 0.5$. The scale $\psi_0$ is $1$ for the damped pendulum and $10$ for Lorenz and Duffing (matching the empirical state-space scale). A jitter $10^{-6}\, I$ is added to the MAP $\Sigma_g$ for numerical positive-definiteness. The same priors are used by the Taylor variant of Appendix~E. Explicit MAP formulas are in the supplementary derivations.

The Koopman matrix $K_g$ is updated by responsibility-weighted least squares in the lifted space: find the matrix that best maps each lifted current state to the corresponding lifted next state, with each transition weighted by its responsibility for cluster $g$. This is the standard EDMD regression generalized to a weighted setting, and it admits a closed-form solution via the Moore--Penrose pseudoinverse:
\[
K_g = \bigl(\tilde{Y}_g W_g \tilde{X}_g^\top\bigr)\bigl(\tilde{X}_g W_g \tilde{X}_g^\top\bigr)^{\dagger},
\]
where $\tilde{X}_g$ and $\tilde{Y}_g$ collect the lifted current and next states as columns, $W_g$ is a diagonal matrix of responsibilities, and $(\cdot)^{\dagger}$ denotes the pseudoinverse. In practice we compute this via SVD, which handles rank-deficient cases gracefully through truncation. The noise variance $\sigma_g^2$ is updated from the responsibility-weighted mean squared residual.

\paragraph{Empty-cluster pruning.} As EM iterates, some clusters may accumulate negligible total responsibility across all training transitions, meaning no region of phase space is better explained by that cluster than by its neighbors. Rather than carrying such clusters through to convergence, we place a weak Dirichlet prior $\mathrm{Dir}(\alpha)$ with $\alpha < 1$ on the mixture weights $\boldsymbol{\pi}$. This penalizes small weights and allows clusters whose total responsibility falls below a threshold to be pruned between iterations, reducing the effective $G$ adaptively. The result is that the final number of clusters reflects the complexity of the data rather than the initial $G$.

\paragraph{Initialization.} EM is sensitive to initialization because the responsibility-weighted objective has many local minima, particularly on systems with strong nonlinearity or wide phase-space coverage. We initialize cluster centers $c_g$ by $k$-means on the training states, which gives a geometrically reasonable starting partition. Each cluster's Koopman matrix $K_g$ is then initialized by running ordinary EDMD on the subset of training transitions initially assigned to that cluster, so the starting per-cluster operators are already locally meaningful rather than random. To reduce sensitivity to the initial partition, EM is run from multiple random restarts and the run achieving the highest log-likelihood is retained.

\paragraph{Rollout.} At inference we have only the current state $x_t$ and no access to the next state, so the residual factor used during training cannot be evaluated. Cluster selection therefore reverts to geometric proximity alone: the active cluster is $g^*(x_t) = \arg\max_g\, \pi_g\, \mathcal{N}(x_t; c_g, \Sigma_g)$, the component whose geometric Gaussian assigns the highest weighted density to the current state. The predicted next state is then
\[
x_{t+1} = c_{g^*} + P_d\, K_{g^*}\, \Phi(x_t - c_{g^*}),
\]
and multi-step rollouts iterate this map, selecting the active cluster afresh at each step. This means the rollout can switch clusters as the trajectory moves through phase space, with the partition boundaries acting as soft regime boundaries.

\section*{Appendix B: Full experimental setup}

\paragraph{Systems.} The three classical systems used in this work, all expressed as continuous-time autonomous ODEs and then integrated to produce discrete-time pairs $(x_t, x_{t+1})$ at fixed step $\Delta t$:

\emph{Lorenz attractor} ($d=3$, polynomial RHS of degree 2; the bilinear terms $xy, xz$ are the highest-degree nonlinearity):
\[
\dot{x} = \sigma(y - x), \quad
\dot{y} = x(\rho - z) - y, \quad
\dot{z} = x y - \beta z,
\]
with parameters $(\sigma, \rho, \beta) = (10, 28, 8/3)$. The attractor exhibits strange-attractor chaos with two folding wings around the unstable origin.

\emph{Damped pendulum} ($d=2$, non-polynomial RHS):
\[
\dot{\theta} = \omega, \qquad
\dot{\omega} = -\sin\theta - \gamma\, \omega,
\]
with damping coefficient $\gamma = 0.2$. The non-polynomial $\sin\theta$ term is the reason no global polynomial lift is exact at any finite degree; this is the system on which the parameter-efficiency advantage of CW-EDMD is largest.

\emph{Double-well Duffing oscillator} ($d=2$, polynomial RHS of degree 3):
\[
\dot{x} = v, \qquad
\dot{v} = x - x^3 - \delta\, v,
\]
with damping $\delta = 0.25$. The cubic restoring force $x - x^3$ creates two stable foci at $x = \pm 1$ and an unstable saddle at $x = 0$. The RHS polynomial degree is 3; in our experiments the EDMD lift degree $q{=}5$ is where global EDMD saturates, with $q{=}3$ matching the RHS degree itself.

\paragraph{Discretization.} All systems are integrated with vectorized RK4. Pair generation produces $(x_t, x_{t+1})$ at fixed step $\Delta t$, with $\Delta t \in \{0.005, 0.01, 0.05, 0.1\}$ across configurations.

\paragraph{Sampling distributions.} Five distributions are swept independently per system: \emph{uniform} over a configurable box, \emph{Gaussian} centered at the origin, \emph{Gaussian mixture} centered at attractor foci, \emph{periodic-noise} (sinusoidal with additive noise), and \emph{trajectory ensemble} (random ICs integrated for several steps, all trajectory points used as training). Additional Lorenz-specific \emph{single-trajectory attractor} sampling is used for two attractor-baseline configurations.

\paragraph{Configurations.} Each system uses 12 configurations: one baseline plus eleven single-axis variations. Per-system baseline values are listed in Table~\ref{tab:baselines}; the eleven variations are enumerated in Table~\ref{tab:configs} as deltas relative to the per-system baseline.

\begin{table}[H]
\centering\footnotesize
\caption{Per-system baseline values. Each non-baseline configuration in Table~\ref{tab:configs} varies one parameter from these reference settings.}
\label{tab:baselines}
\begin{tabular}{l l l l}
\toprule
Parameter                          & Pendulum                              & Duffing                    & Lorenz \\
\midrule
training pairs $N$                 & $4000$                                & $4000$                     & $4000$ \\
integrator step $\Delta t$         & $0.05$                                & $0.05$                     & $0.01$ \\
EM iterations $n_{\text{iter}}$    & $100$                                 & $80$                       & $100$ \\
EM restarts $n_{\text{restarts}}$  & $2$                                   & $2$                        & $2$ \\
sampling distribution              & uniform on $[-\pi,\pi]\times[-3,3]$  & uniform on $[-2,2]^2$      & single trajectory on attractor \\
\bottomrule
\end{tabular}
\end{table}

\begin{table}[H]
\centering\footnotesize
\caption{The eleven non-baseline configurations per system, each varying a single parameter from the baseline (Table~\ref{tab:baselines}).}
\label{tab:configs}
\begin{tabular}{l l l}
\toprule
Configuration                & Factor varied             & Value (relative to baseline) \\
\midrule
small training data          & $N$                       & $500$ ($\nicefrac{1}{8}{\times}$ baseline) \\
large training data          & $N$                       & $16{,}000$ ($4{\times}$ baseline) \\
short integrator step        & $\Delta t$                & $\nicefrac{1}{2}{\times}$ baseline \\
long integrator step         & $\Delta t$                & $2{\times}$ baseline \\
narrow box                   & spatial domain            & tighter box, system-specific \\
wide box                     & spatial domain            & wider box, system-specific \\
heavy fit budget             & EM iterations / restarts  & $n_{\text{iter}}{\geq}240,\ n_{\text{restarts}}{=}5$ \\
Gaussian sampling            & sampling distribution     & Gaussian centered at origin \\
Gaussian-mixture sampling    & sampling distribution     & Gaussian mixture at attractor foci \\
periodic-noise sampling      & sampling distribution     & sinusoidal $+$ additive noise \\
trajectory-ensemble sampling & sampling distribution     & random ICs integrated forward \\
\bottomrule
\end{tabular}
\end{table}

\paragraph{Seeds.} 10 fixed seeds per (system, configuration, method): $\{1, 42, 101, 307, 1001, 7789, 13245, 11, 103, 13\}$. Seeds control train/test sampling, EM initialization, and integrator noise.

\paragraph{Methods.} EDMD and CW-EDMD are evaluated at matched polynomial lift degrees per system: $\{2,3\}$ for Lorenz, $\{2,4\}$ for the damped pendulum, $\{2,3,4,5\}$ for Duffing. CW-EDMD additionally sweeps cluster count $G$: $\{2,4,8,16\}$ for pendulum and Duffing; $\{5,12,20,50\}$ for Lorenz (scaled up because the attractor's larger state-space coverage benefits from finer partitioning).

\paragraph{Metrics.} (i)~\emph{One-step error}: mean $\ell_2$ prediction error on the held-out test set. (ii)~\emph{Rollout error at $H$ seconds}: mean $\ell_2$ error of the iterated map at horizons $H \in \{1,2,5,10,20\}$\,s, averaged over test initial conditions.

\paragraph{Statistics.} For each (system, configuration, metric) we run all methods over 10 seeds and compute mean and 95\% CI. \emph{Paired Wilcoxon signed-rank tests} (per-seed pairing) test the matched-degree comparison ``CW-EDMD $q$ vs.\ EDMD $q$''. A cross-configuration outcome is a \emph{win} if \textit{p}\,$<\,0.05$ and the CW-EDMD mean is lower; a \emph{loss} if \textit{p}\,$<\,0.05$ and higher; a \emph{tie} otherwise. We report wins / losses / ties tallied across the 12 configurations.

\section*{Appendix C: Per-system detailed results}

\begin{figure}[H]
	\begin{center}
		\includegraphics[width=0.85\textwidth]{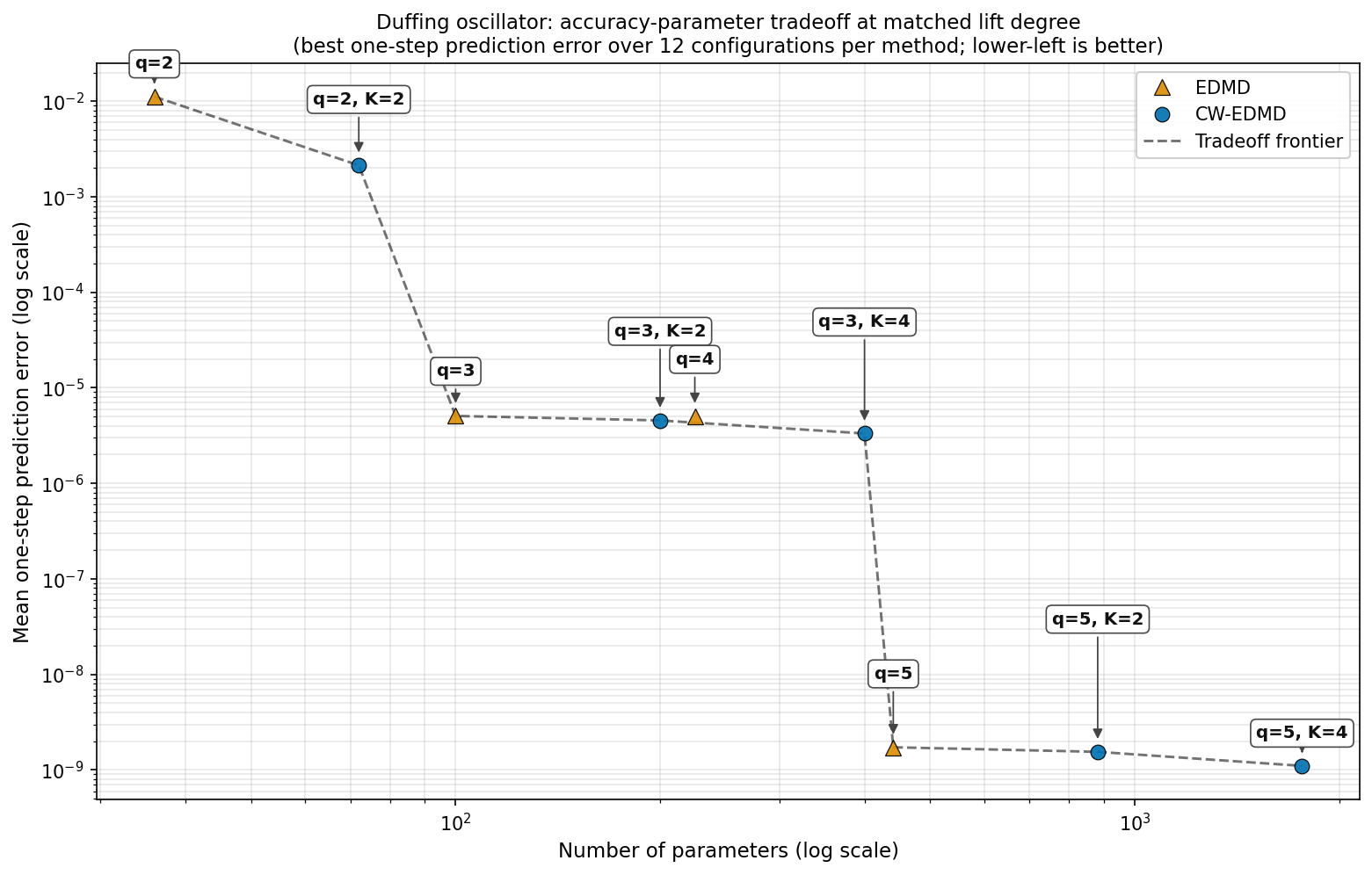}
	\end{center}
	\caption{\textbf{Duffing: accuracy-parameter tradeoff at matched lift degree.} On the polynomial-RHS Duffing system, CW-EDMD's matched-$q$ partitioning offers a small absolute improvement over global EDMD at significantly higher parameter cost; at $q{=}5$, both methods saturate and the tradeoff narrows further.}
	\label{fig:pareto-duffing}
\end{figure}

\begin{figure}[H]
	\begin{center}
		\includegraphics[width=0.85\textwidth]{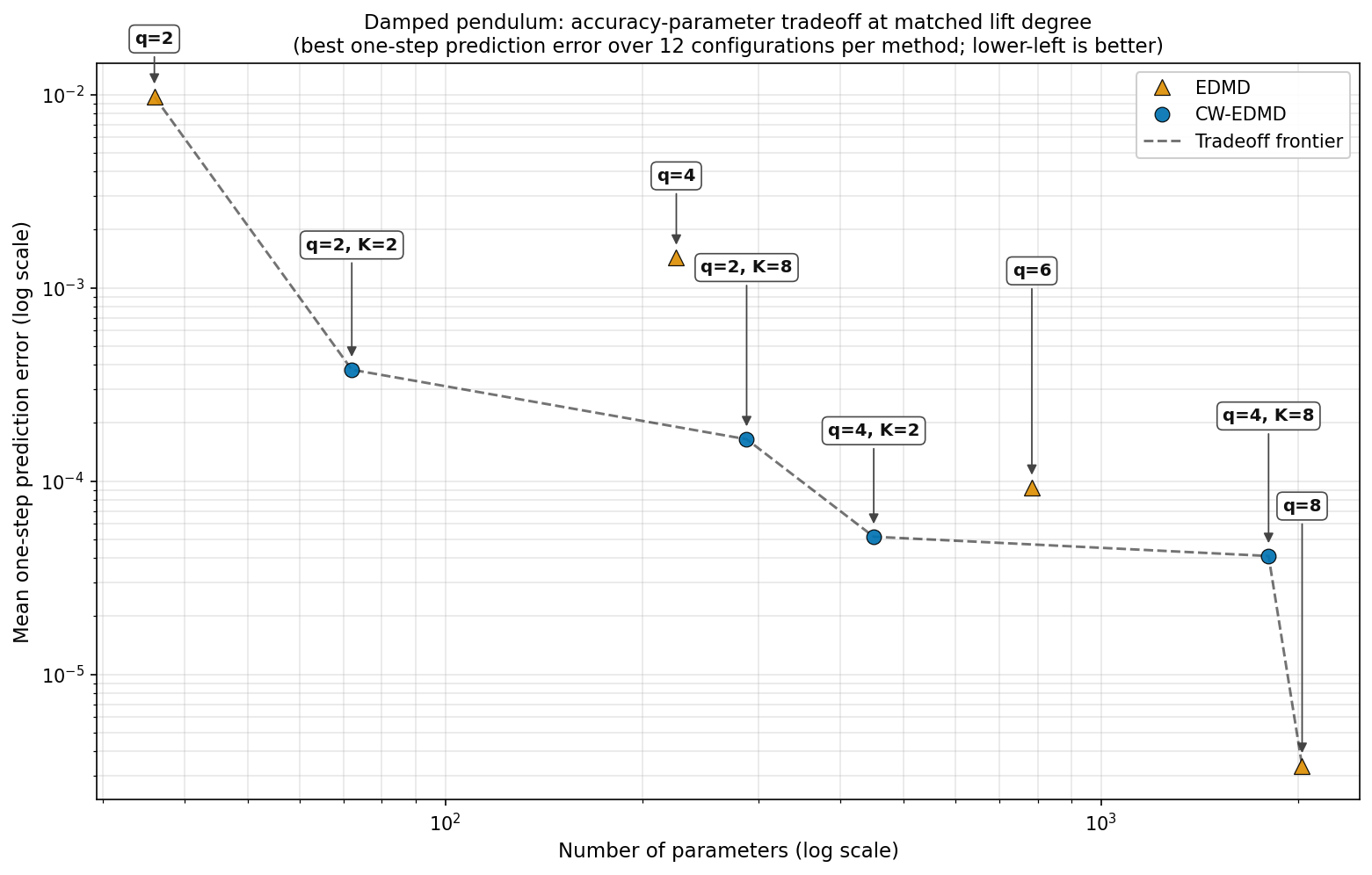}
	\end{center}
	\caption{\textbf{Damped pendulum: accuracy-parameter tradeoff.} Two scaling axes are visible. At matched lift degree, CW-EDMD's partitioning yields a large multiplicative gain (roughly two orders of magnitude at $q{=}4, G{=}16$) for $G$-fold parameter cost. Independently, scaling the global lift to $q{=}6, 8$ also reduces EDMD error substantially; on this $d{=}2$ system $M_q = \binom{d+q}{q}$ stays small ($M_8 = 45$) so high-$q$ EDMD is operationally viable and reaches the lowest absolute error in the figure. The matched-$q$ comparison isolates the partitioning effect; the unmatched comparison shown here exposes the lift-scaling alternative, discussed in the Unmatched-degree paragraph below.}
	\label{fig:pareto-pendulum}
\end{figure}

\begin{figure}[H]
	\begin{center}
		\includegraphics[width=0.85\textwidth]{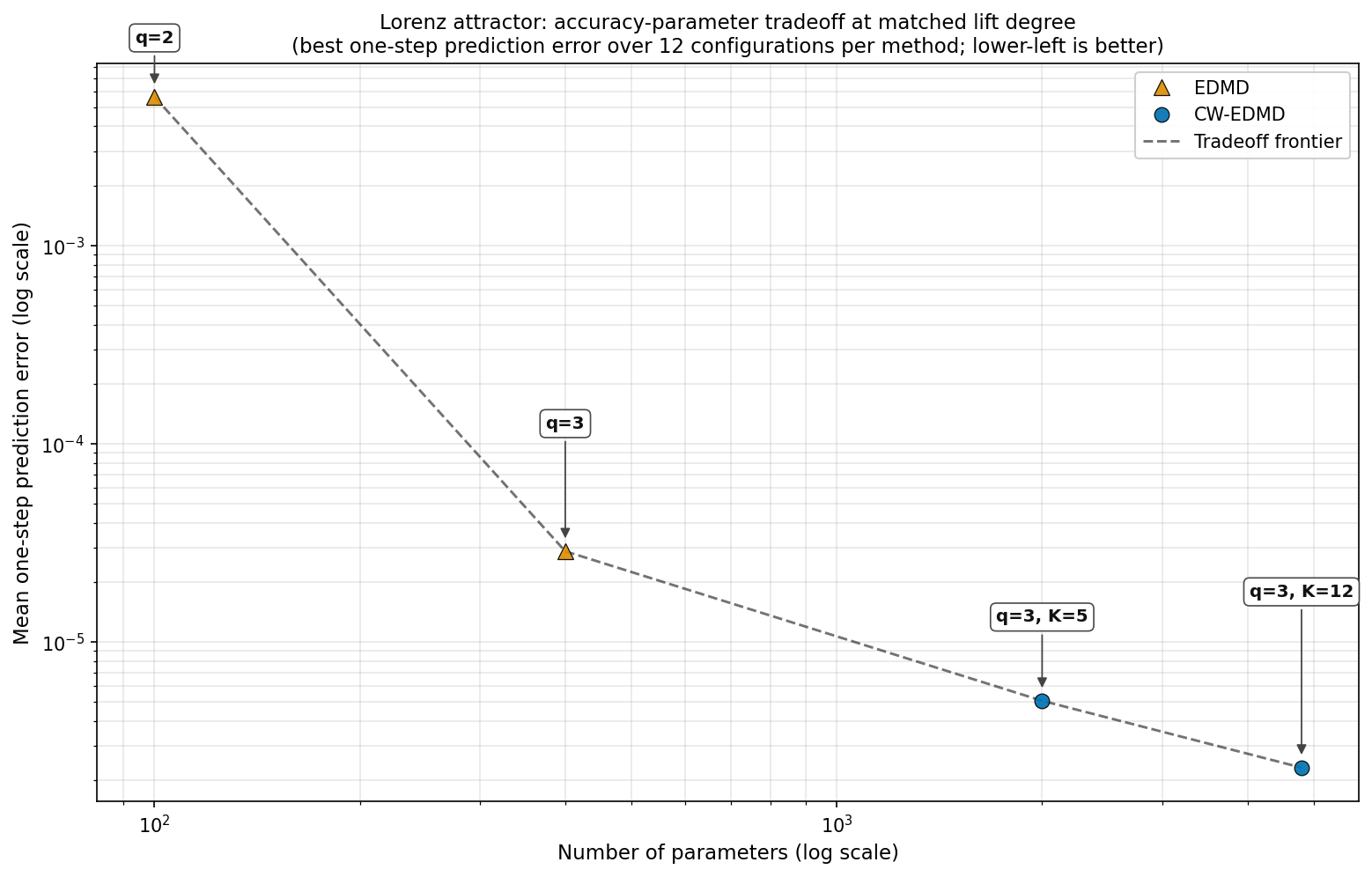}
	\end{center}
	\caption{\textbf{Lorenz: accuracy-parameter tradeoff at matched lift degree.} At the matched lift $q{=}3$, CW-EDMD with $G{=}12$ achieves an order-of-magnitude error reduction over global EDMD at $12\times$ the parameter cost. The single configuration in which CW-EDMD loses to EDMD is the small-training-data configuration ($N{=}500$), where each cluster receives roughly $40$ samples to fit a $400$-parameter Koopman matrix; the two configurations in which CW-EDMD ties EDMD on the 5\,s rollout are this same configuration and the short-integrator-step configuration ($\Delta t{=}0.005$).}
	\label{fig:pareto-lorenz}
\end{figure}

\paragraph{Tradeoff frontiers (Figures \ref{fig:pareto-duffing}--\ref{fig:pareto-lorenz}).} Each figure plots accuracy against parameter count at matched lift degree; CW-EDMD is not a Pareto-dominance claim over global EDMD but a tradeoff curve: at fixed lift degree $q$, $G$-fold partitioning buys substantially lower error for $G$-fold more parameters. Shared conventions across the three panels: each marker is a Pareto-frontier configuration (one method at its best-case setting over $12$ configs $\times$ $10$ seeds); orange~$\blacktriangle$~= EDMD at polynomial degree $q$; blue~$\bullet$~= CW-EDMD (the focus of this paper) at degree $q$ and cluster count $G$; each marker is labeled by its configuration; dashed line~= tradeoff frontier; lower-left is better. Ablation variants (GMM-EDMD, CW-Taylor, GMM-Taylor) are reported quantitatively in Table~\ref{tab:within-edmd-regimes} and Appendix~D rather than overlaid on these figures.

\paragraph{Damped pendulum.} At matched lift degree $q{=}4$, median prediction error in dimensionless state-space units across the 12 configurations and 10 seeds is reported in Table~\ref{tab:pendulum}. Median is reported in preference to mean because rollout error on a small subset of high-dynamic-range configurations diverges to large values that dominate the cross-configuration mean.
\begin{table}[H]
\caption{Damped pendulum at matched lift degree $q{=}4$: median prediction error (dimensionless state-space units) across 12 configurations and 10 seeds. EDMD $q{=}2$ is shown as a low-lift baseline; EDMD $q{=}4$ is the matched-degree baseline for the CW-EDMD rows. Bold marks the column-wise minimum.}
\label{tab:pendulum}
\centering\small
\begin{tabular}{lrrr}
\toprule
Method & Params & one-step & 5\,s \\
\midrule
EDMD $q{=}2$ (low lift) & 36 & $2.5\!\times\!10^{-2}$ & $1.1\!\times\!10^{0}$ \\
EDMD $q{=}4$ (matched) & 225 & $4.6\!\times\!10^{-3}$ & $3.1\!\times\!10^{-1}$ \\
\midrule
CW-EDMD $q{=}4, G{=}4$ & 900 & $8.5\!\times\!10^{-5}$ & $5.0\!\times\!10^{-3}$ \\
CW-EDMD $q{=}4, G{=}8$ & 1800 & $6.1\!\times\!10^{-5}$ & $4.1\!\times\!10^{-3}$ \\
CW-EDMD $q{=}4, G{=}16$ & 3600 & $\mathbf{5.6\!\times\!10^{-5}}$ & $\mathbf{2.5\!\times\!10^{-3}}$ \\
\bottomrule
\end{tabular}

\end{table}
\noindent At matched lift degree $q{=}4$, CW-EDMD $G{=}16$ wins on $11/12$ configurations on one-step and $11/12$ on 5\,s rollout against EDMD $q{=}4$. The single tied configuration on each metric is the periodic-noise sampling distribution, where the sinusoidal-noise training distribution drives both methods to comparable error (ratio EDMD/CW-EDMD $\approx 1$). The intermediate $G{=}4, 8$ rows show the same qualitative win pattern at higher mean error, illustrating that the partitioning gain compounds as $G$ grows on this non-polynomial-RHS system.

\paragraph{Unmatched-degree comparison on pendulum.} Figure~\ref{fig:pareto-pendulum} also exposes EDMD at lift degrees $q{=}6, 8$, beyond the matched-degree pair reported in Table~\ref{tab:pendulum}. At $q{=}8$, global EDMD reaches one-step error of order $3\!\times\!10^{-6}$ with $M_q^2 = 45^2 \approx 2000$ parameters, lower than any CW-EDMD configuration we ran on this system. Two clarifications. First, this is consistent with the matched-$q$ claim: at the same lift the per-$q$ rows of Table~\ref{tab:pendulum} still show CW-EDMD beating EDMD by roughly two orders of magnitude. Second, the lift-scaling alternative is only operationally accessible on low-dimensional smooth-RHS systems. The per-cluster parameter count is $M_q^2 = \binom{d+q}{q}^2$ in state dimension $d$: for pendulum ($d{=}2$), $M_8 = 45$; for Lorenz ($d{=}3$), $M_8 = 165$; for $d{=}4$, $M_8 = 495$; for $d{=}5$, $M_8 = 1287$. The Lorenz and Duffing sweeps (Tables~\ref{tab:lorenz}, \ref{tab:duffing}) consequently stop at $q{=}3$ and $q{=}5$ respectively, where global EDMD already saturates and scaling $q$ further is neither necessary nor cheap. The pendulum is the unique system in the corpus where the lift-scaling axis is unconstrained, and the figure now shows it. This is also consistent with the two-mechanism decomposition in Appendix~D: at $q{=}8$ the polynomial basis largely captures $\sin\theta$ on the sampled domain, so the \emph{mismatched-lift} regime in Table~\ref{tab:within-edmd-regimes} (where residual-aware partitioning dominates) no longer applies, and the comparison shifts to a parameter-efficiency tradeoff between two ways of spending capacity.

\paragraph{Duffing.} At three matched polynomial degrees, median prediction error across the 12 configurations and 10 seeds is reported in Table~\ref{tab:duffing}, with EDMD baselines at every matched degree for direct comparison.
\begin{table}[H]
\caption{Duffing oscillator: median prediction error (dimensionless state-space units) across 12 configurations and 10 seeds. Top block: EDMD at four polynomial lift degrees ($q{=}2$ as a low-lift reference; $q{=}3, 4, 5$ each as the matched-degree baseline for the corresponding CW-EDMD row). Bottom block: CW-EDMD at three matched lift degrees. Bold marks the column-wise minimum.}
\label{tab:duffing}
\centering\small
\begin{tabular}{lrrr}
\toprule
Method & Params & one-step & 5\,s \\
\midrule
EDMD $q{=}2$ (low lift) & 36 & $5.2\!\times\!10^{-2}$ & $7.7\!\times\!10^{-1}$ \\
EDMD $q{=}3$ (matched) & 100 & $6.5\!\times\!10^{-5}$ & $6.6\!\times\!10^{-3}$ \\
EDMD $q{=}4$ (matched) & 225 & $6.5\!\times\!10^{-5}$ & $6.6\!\times\!10^{-3}$ \\
EDMD $q{=}5$ (matched) & 441 & $6.8\!\times\!10^{-8}$ & $4.6\!\times\!10^{-6}$ \\
\midrule
CW-EDMD $q{=}3, G{=}8$ & 800 & $2.4\!\times\!10^{-5}$ & $2.7\!\times\!10^{-3}$ \\
CW-EDMD $q{=}4, G{=}4$ & 900 & $1.7\!\times\!10^{-5}$ & $1.3\!\times\!10^{-3}$ \\
CW-EDMD $q{=}5, G{=}4$ & 1764 & $\mathbf{4.1\!\times\!10^{-8}}$ & $\mathbf{3.1\!\times\!10^{-6}}$ \\
\bottomrule
\end{tabular}

\end{table}
\noindent CW-EDMD wins on 35 of 36 cells across the three matched degrees against EDMD (paired Wilcoxon, $p<0.05$); the single tie is the periodic-noise sampling configuration at $q{=}5$, where the Wilcoxon test does not separate the methods. The within-Taylor ablation against the standard GMM-clustered local model contrast point of the CWM literature is reported in Appendix~D.

\paragraph{Lorenz.} Median prediction error across the 12 configurations and 10 seeds is reported in Table~\ref{tab:lorenz}. The metric \emph{r5s} reaches the rollout-step cap on the chaotic configurations and saturates; \emph{one-step} is the cleanest cross-method comparison.
\begin{table}[H]
\caption{Lorenz attractor: median prediction error (dimensionless state-space units) across 12 configurations and 10 seeds. Top block: EDMD at two polynomial lift degrees. Middle block: CW-EDMD at lift degree $q{=}2$ (mismatched). Bottom block: CW-EDMD at the matched lift degree $q{=}3$. Bold marks the column-wise minimum.}
\label{tab:lorenz}
\centering\small
\begin{tabular}{lrrr}
\toprule
Method & Params & one-step & 5\,s \\
\midrule
EDMD $q{=}2$ & 100 & $3.8\!\times\!10^{-2}$ & $1.5\!\times\!10^{1}$ \\
EDMD $q{=}3$ & 400 & $1.6\!\times\!10^{-3}$ & $2.4\!\times\!10^{0}$ \\
\midrule
CW-EDMD $q{=}2, G{=}5$ & 500 & $3.3\!\times\!10^{-3}$ & $7.0\!\times\!10^{0}$ \\
CW-EDMD $q{=}2, G{=}12$ & 1200 & $6.4\!\times\!10^{-4}$ & $1.4\!\times\!10^{0}$ \\
CW-EDMD $q{=}2, G{=}20$ & 2000 & $4.6\!\times\!10^{-4}$ & $7.7\!\times\!10^{-1}$ \\
\midrule
CW-EDMD $q{=}3, G{=}5$ & 2000 & $1.7\!\times\!10^{-4}$ & $1.3\!\times\!10^{-1}$ \\
CW-EDMD $q{=}3, G{=}12$ & 4800 & $\mathbf{1.6\!\times\!10^{-4}}$ & $1.1\!\times\!10^{-1}$ \\
CW-EDMD $q{=}3, G{=}20$ & 8000 & $1.7\!\times\!10^{-4}$ & $\mathbf{9.4\!\times\!10^{-2}}$ \\
\bottomrule
\end{tabular}

\end{table}
\noindent At the matched lift degree $q{=}3$, CW-EDMD with $G{=}12$ records $11$ wins and $1$ loss on one-step and $10$ wins and $2$ ties on the 5\,s rollout against EDMD $q{=}3$ (paired Wilcoxon, $p < 0.05$). The single loss is the small-training-data Lorenz configuration ($N{=}500$ pairs, $8\times$ smaller than the $N{=}4000$ baseline), where each cluster's $M_q^2 = 400$-parameter Koopman operator at $q{=}3$ receives roughly $N/G \approx 42$ training samples after responsibility assignment, an order-of-magnitude underdetermination per cluster. The two 5\,s ties are this same small-training-data configuration and the short-integrator-step configuration ($\Delta t{=}0.005$, half the baseline); the latter is not a per-cluster scarcity case ($N/G \approx 333$, identical to the winning baseline) and the tie reflects high temporal correlation between consecutive samples on the chaotic attractor reducing the effective decorrelated sample count. At fixed $G{=}5$ ($500$--$2000$ parameters), increasing the lift degree from $q{=}2$ to $q{=}3$ decreases one-step error by roughly two orders of magnitude, demonstrating that the matched lift degree is essential for the partitioning advantage on this polynomial-RHS system. The mismatched $q{=}2$ CW-EDMD entries are included for ablation.

\section*{Appendix D: Generality and initialization}

\paragraph{CWM framing and approximator-agnostic per-cluster fit.} CW-EDMD instantiates the cluster-weighted-model framework~\citep{gershenfeld1999nature,ingrassia2014linear,punzo2014polynomial} with EDMD as the per-cluster predictor. The CWM joint density $p(x_t, x_{t+1}) = \sum_g \pi_g\, p_g(x_t)\, p_g(x_{t+1}\mid x_t)$ is agnostic in the per-cluster predictor $p_g(x_{t+1}\mid x_t)$. We provide a Taylor-expansion variant in our implementation that replaces the per-cluster EDMD operator with a first-order centered linearization $\hat{x}_{t+1\mid g} = x_t + \Delta t\,[f(c_g) + J(c_g)(x_t - c_g)]$ (one explicit-Euler step on the linearized vector field) when analytic $f$ and $J$ are available at cluster centers; the same EM machinery applies. This drop-in substitutability is what we mean by \emph{model-agnostic}: any local approximator that admits a fit-time loss and an inference-time prediction can be substituted. The headline matched-degree comparisons in the main text are restricted to EDMD by design, to make the partitioning hypothesis a clean controlled test on one approximator family.

\paragraph{Within-EDMD ablation: CW-EDMD vs.\ GMM-clustered EDMD at matched $(q, G)$.} To isolate the contribution of the residual-aware responsibility update from the partitioning effect itself, we compare CW-EDMD against an ablation in which the residual factor is removed from the E-step (the responsibility update in Appendix~A, \emph{EM updates}). Concretely, the CW-EDMD responsibility update
\[
r_{ig} \propto \pi_g\, \mathcal{N}(x_t^{(i)}; c_g, \Sigma_g)\, \mathcal{N}(\Delta x_g^{(i)}; 0, \sigma_g^2 I)
\]
is replaced by the geometry-only GMM-style update
\[
r_{ig} \propto \pi_g\, \mathcal{N}(x_t^{(i)}; c_g, \Sigma_g),
\]
which is the $\sigma_g^2 \to \infty$ limit of the residual factor (the residual likelihood degenerates to a constant and drops out). The M-step (updates to $c_g$, $\Sigma_g$, $K_g$, $\pi_g$) is held identical, and the per-cluster EDMD predictor and lift degree $q$ are held fixed; the \emph{only} difference between the two variants is the presence or absence of the residual factor in the E-step. We denote this variant \emph{GMM-EDMD}. It is not a standard baseline in the EDMD literature; it is the natural within-family ablation that isolates the residual-aware responsibility mechanism. The ablation is run on all three systems at every configured $(q, G)$ pair, with paired Wilcoxon tests per configuration. Across $324$ paired comparisons (system $\times$ $q$ $\times$ $G$ $\times$ configuration), CW-EDMD wins on $271$ ($84\%$) at $p<0.05$ with lower mean error, ties on $49$ ($15\%$), and loses on $4$ ($1\%$).

The outcome decomposes by lift regime (Table~\ref{tab:within-edmd-regimes}). When the polynomial lift cannot fully capture the local dynamics, either because the RHS is non-polynomial (damped pendulum) or because the lift is under-exact for a polynomial RHS (Lorenz at $q{=}2$, Duffing at $q{=}2$), the residual-aware E-step contributes decisively and CW-EDMD wins on nearly every configuration. When the lift is sufficient and global EDMD itself saturates, CW-EDMD still wins on the majority of configurations, but the residual-aware margin shrinks: ties concentrate at the specific $(q, G)$ pairs where both methods reach the analytical floor and the Wilcoxon test cannot separate them. The losses concentrate in the per-cluster-scarcity regime where the per-cluster sample count $N/G$ is small relative to the per-cluster parameter count $M_q^2$: an under-determined $K_g$ produces residuals dominated by fitting noise, and CW-EDMD's residual-aware E-step trusts that noise rather than partition signal, while GMM-EDMD's geometry-only E-step ignores the residuals and is more robust in this regime. The residual-aware update is therefore a feature only when the residuals it consumes carry usable signal. This is the direct mechanistic measurement that closes the inference: residual-awareness is the source of CW-EDMD's gain in regimes where the partition has work to do, and partitioning carries the gain in regimes where the lift suffices.

\begin{table}[H]
\centering\footnotesize
\caption{Within-EDMD ablation (CW-EDMD vs.\ GMM-EDMD) decomposed by lift regime, paired Wilcoxon wins / losses / ties. \emph{Mismatched}: non-polynomial RHS (pendulum) or polynomial RHS at under-exact lift; the lift cannot capture local dynamics. \emph{Sufficient}: polynomial RHS at the degree where global EDMD itself saturates; CW-EDMD still wins on the majority of configurations, with ties concentrated at the saturated $(q, G)$ pairs where both methods reach the analytical floor and the Wilcoxon test cannot separate them. \emph{Per-cluster scarcity}: per-cluster sample count $N/G$ is small relative to per-cluster parameter count $M_q^2$ (Lorenz $q{=}3$, $G{=}50$: $N/G = 80$ against $M_q^2 = 400$).}
\label{tab:within-edmd-regimes}
\begin{tabular}{llc}
\toprule
Lift regime  & Slice                                       & W/L/T \\
\midrule
Mismatched   & Pendulum; Lorenz $q{=}2$; Duffing $q{=}2$    & $184/1/7$ \\
Sufficient   & Duffing $q{\geq}3$; Lorenz $q{=}3$ (low $G$) & $87/0/42$ \\
Per-cluster scarcity  & Lorenz $q{=}3, G{=}50$                & $0/3/0$ \\
\midrule
Total        &                                              & $271/4/49$ \\
\bottomrule
\end{tabular}
\end{table}

\paragraph{Two-mechanism decomposition of CW-EDMD's advantage.} The within-EDMD ablation also implies a decomposition of CW-EDMD's gain over global EDMD into two contributing mechanisms: (i)~\emph{partitioning}, measurable as GMM-EDMD vs.\ EDMD (a per-cluster EDMD operator on a geometrically-clustered partition beats a single global operator at matched lift), and (ii)~\emph{residual-aware responsibilities}, measurable as CW-EDMD vs.\ GMM-EDMD. On polynomial-RHS systems at the lift where EDMD saturates (Lorenz $q{=}3$, Duffing $q{\geq}3$), mechanism (i) carries the gain: partitioning alone already closes most of the gap, and the residual-aware update contributes negligibly. On the damped pendulum at any $q$ and on polynomial-RHS systems at lower lift, mechanism (ii) dominates: geometry-only partitioning is barely better than global EDMD, and the residual-aware E-step delivers the bulk of CW-EDMD's improvement.

\paragraph{Cross-predictor confirmation: within-Taylor ablation.} The same ablation in the Taylor predictor branch (CW-Taylor vs.\ GMM-Taylor, the classical CWM-vs-two-stage contrast point in the statistical-clustering literature~\citep{ingrassia2014linear}) shows the same qualitative pattern, with the residual-aware mechanism contributing decisively at moderate-to-large $G$ on all three systems. The agreement across predictor families confirms the mechanism is intrinsic to the responsibility-update form, not specific to the EDMD predictor.

\paragraph{Initialization.} We initialize $c_g$ by $k$-means and $K_g$ by ordinary EDMD on the points initially assigned to cluster $g$, with multi-restart EM. Multi-restart is essential on Duffing wide-box configurations where the cubic nonlinearity creates deep local minima.

\paragraph{Computational cost.} EDMD fits a single closed-form regression of size $M_q^2$ in one shot. CW-EDMD replaces this with $G$ regressions of the same size per EM iteration, with $T$ iterations and $R$ restarts, giving a fit cost of order $R \cdot T \cdot G \cdot M_q^2 \cdot N$ versus $M_q^2 \cdot N$ for EDMD. In the configurations reported here the worst-case CW-EDMD fit is roughly three orders of magnitude more expensive than a single EDMD fit. Inference cost is comparable to EDMD: rollout requires one $K_g\,\Phi(\cdot)$ matrix-vector product per step plus a Gaussian-likelihood evaluation to pick the active cluster.

\paragraph{Limitations.} Four limitations of the present study are worth flagging. First, we do not provide a quantitative predictive threshold for when CW-EDMD breaks down; the only failure observed in the corpus is the extreme per-cluster-scarcity Lorenz configuration ($N{=}500$, $N/G \approx 42$ samples per cluster against $M_q^2 = 400$ parameters per cluster), and a quantitative rule that maps $(N, G, M_q, \Delta t, \text{system})$ to a predicted win/loss outcome remains future work. Second, the Lorenz cross-system claim is empirically weaker than the Duffing one: on Lorenz we cannot claim CW-EDMD strictly dominates EDMD on every configuration at the matched lift, whereas on Duffing we can. The single loss and the two ties are mechanistically explained (per-cluster scarcity, temporal correlation; Appendix~C), and the explanations are load-bearing: they tell the reader where the method's empirical dominance frays. We flag this as a caveat to the cross-system claim, not as a method-level failure. Third, the corpus is restricted to autonomous low-dimensional systems with smooth dynamics; control-input-driven systems and higher-dimensional flows (where CW-EDMD's parameter count grows quadratically in $M_q$) are out of scope here. Fourth, on the lowest-dimensional system in the corpus (pendulum, $d{=}2$) global EDMD remains operationally feasible at high lift ($q{=}8$, $M_8 = 45$) and reaches lower one-step error than any CW-EDMD configuration we ran (Figure~\ref{fig:pareto-pendulum}); the matched-$q$ framing of the headline comparison therefore flatters CW-EDMD on this system, because the alternative scaling axis is unusually cheap here. On Lorenz and Duffing the lift-scaling alternative is constrained either by saturation (the $q$ at which global EDMD reaches the analytical floor) or by combinatorial blow-up of $M_q$, and the matched-$q$ comparison is the operationally-relevant one.

\section*{Appendix E: Reproducibility}

All experimental configurations are versioned as YAML files dispatched through a single statistical-validation driver (\texttt{validation/run\_statistical.py --config <name>.yaml}). Per-seed JSON results are written for every (system, configuration, method, seed) combination and aggregated post-hoc into the long-form CSV used for the figures and tables in this paper. The full corpus (39{,}564 observations across systems / configurations / methods / seeds / metrics) and code are available at \url{https://github.com/agencyenterprise/cluster_weighted_edmd}.

\end{document}